# Bayesian Prediction for Artificial Intelligence

Matthew Self     Peter Cheeseman

NASA Ames Research Center
Mail Stop 244-17
Moffett Field, CA 94035

## Abstract

This paper shows that the common method used for making predictions under uncertainty in AI and science is in error. This method is to use currently available data to select the best model from a given class of models—this process is called *abduction*—and then to use this model to make predictions about future data. The correct method requires averaging over all the models to make a prediction—we call this method *transduction*. Using transduction, an AI system will not give misleading results when basing predictions on small amounts of data, when no model is clearly best. For common classes of models we show that the optimal solution can be given in closed form.

## 1 Introduction

This paper is concerned with Bayesian probabilistic prediction for Artificial Intelligence (AI). Most text books on Bayesian analysis are concerned with Bayesian decision making and only make passing reference to the prediction problem. Current statistical methods for making predictions based on previous data work by using the previous data to choose the best model, called *abduction*, and then use this model to make predictions. This method is in error—the correct method requires taking all the possible models into account with suitable weighting, a method we call *transduction*. Although the two methods produce the same results for asymptotically large initial samples, this paper shows that there can be considerable differences in the answers produced by transduction and abduction methods. For many common models, transduction can be performed in closed form. This paper presents Bayesian prediction in an automated reasoning context and shows how these results compare with standard conceptions of the scientific method and philosophy. Although most of the following material can be found in one form or another in texts on probability theory and statistics, our emphasis is on prediction in AI systems, such as expert systems.

In many expert systems, the expert provides a model, which allows the expert system to make predictions. For example, a nuclear power plant controller can determine the probability of various malfunctions given data from various sensors. The system might use the most likely malfunction model to predict the consequences of possible interventions. This paper argues that this approach is wrong if the aim is to make the best predictions. The problem is that the prediction is not based on a weighted combination of all the possible failure models, but only on one. We show that using transduction guarantees that AI systems will produce appropriate predictions regardless of how much data the prediction is based on. The abduction method could produce misleading results without the system being aware of it.

## 2 Bayesian Prediction

In a Bayesian framework, prediction corresponds to providing a probability distribution over possible future events. This is to be contrasted with only determining the most probable future event, or determining some other *point estimate*, such as the mean value of a variable, or logical approaches where it is assumed that the world can be modeled without error. Having obtained the probability distribution of future data, optimal actions can be chosen using standard decision theory with a suitable utility function [e.g. Raiffa]. We are concerned only with optimal prediction, rather than optimal decision making, since one needs the former to perform the latter. In many situations, the predictor does not know what use will be made of the estimates provided. For example, a scientist measuring the yield strength of steel does not know how this value will be used in a decision making context. In such situations, a probability distribution over the possible values should be given, so any future decision maker can supply the appropriate utility function to make the optimal decision. The near universal identification of Bayesian analysis with Bayesian decision making [e.g. Berger, Raiffa, and Schlaifer, etc.], is only an historical accident—Bayesian prediction is also an important problem. If the utility function for the particular problem at hand is known, there may be computational advantages in combining the optimal decision problem with the prediction problem. However, the prediction problem can often be solved once in closed form, leaving any particular decision problem to be calculated as needed. Although the ability of Bayesian analysis to provide the needed predictive distributions is documented in the literature [Berger, Winkler], very little emphasis is placed on these techniques.

## 3 Theory

Here we present only a brief summary of Bayesian theory. We assume that the reader is already reasonably familiar with the basic concepts. A detailed example follows in the next section.

### 3.1 Deduction

Bayesian *deduction* is inference from models to data. That is, using a known model we wish to predict the probabilities of data given that model. This process is entirely described by the likelihood function,

$$p(x \mid \theta). \qquad (1)$$

This is the probability of observing the (unknown) data, $x$, *given* the (known) model, $\theta$. That is, the probability distribution of the data is a *function* of the model. The data might be the symptoms of a patient, while the model might be a particular disease. In this case the likelihood function predicts the probability of various symptoms of a patient with a known disease. It is important to realise that the data, $x$, are *observable*, while the models, $\theta$, are not. In a Bayesian framework, one defines a model by describing its deductive behaviour through the likelihood function. For each model the likelihood function gives a probability distribution over the data, and thus the likelihood function is the *definition* of the model, as shown in Figure 1

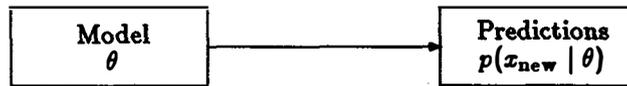

Figure 1: Probabilistic Deduction

### 3.2 Induction

Although models are defined by their deductive behaviour, one often wishes to induce a particular model given observed data. This inverse problem is solved through the use of Bayes' Theorem,

$$p(\theta \mid x) = \frac{p(x \mid \theta)\, p(\theta)}{p(x)} \qquad (2)$$

where $p(\theta)$ is the prior probability of the models, $\theta$, and $p(x)$ is the prior predictive distribution of the data, $x$. If the models, $\theta$, are mutually exclusive and exhaustive, we can express $p(x)$ in terms of the likelihood function and the prior distribution of $\theta$,

$$p(x) = \sum_\theta p(x \mid \theta)\, p(\theta). \qquad (3)$$

If we are only trying to determine the relative probabilities of the models, we need not evaluate the normalising constant, $p(x)$.

For models with adjustable parameters we treat each possible set of values for the parameters as a separate model, and for continuous parameters we use probability density functions rather than probability mass functions, and the summations become integrals. For both the discrete and continuous case, Bayesian induction provides the posterior distribution of the models given the data.

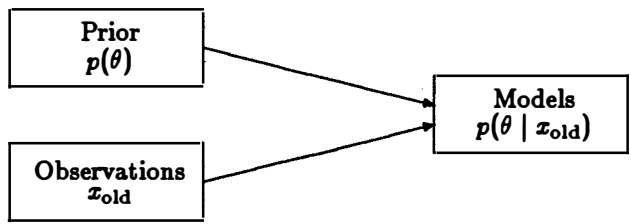

Figure 2: Induction

## 3.3 Transduction

If we now combine these two forms of inference by calculating the distribution of future observations based on past observations we are performing what we call *transduction*. Transduction is the mapping from past data to future data via a model space, as shown in Figure 3. The heart of transduction is the so-called predictive distribution. It is the marginal distribution of the data considering all of the possible models. That is, each model gives a specific prediction about future data, and this prediction is weighted by the probability of that model, to give the combined prediction. We can form both a prior and posterior predictive distribution by weighting the predictions with either the prior distribution of the models, or with the posterior distribution of the models given previous data. The *prior* predictive distribution is just Equation 3, which is also the denominator in Bayes' Theorem. The *posterior* predictive distribution is given by

$$p(x_{\text{new}} \mid x_{\text{old}}) = \sum_{\theta} p(x_{\text{new}} \mid \theta, x_{\text{old}}) \, p(\theta \mid x_{\text{old}}), \qquad (4)$$

and gives the marginal distribution of future data given the past data.

As a simple example of transduction, if a military commander is unsure in which of two locations his enemy is hiding, he must consider the expected results from both alternatives when predicting the consequences of possible deployments of his troops. Even if one of the alternatives is much less likely than the other, if the results of its being true would be devastating, it may dominate the predictions, and hence any decision. To only consider the most probable state of events can lead to disastrous consequences.

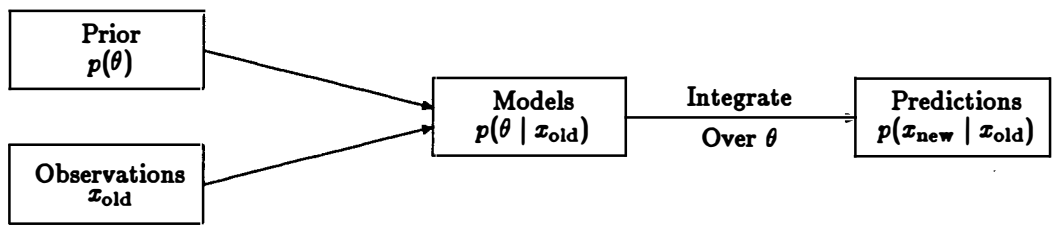

Figure 3: Transduction

By substituting the definition of the posterior distribution of the models from Equation 2 into Equation 4, we can express the posterior predictive distribution solely in terms of the likelihood function and the prior probabilities of the models,

$$p(x_{\text{new}} \mid x_{\text{old}}) = \frac{\sum p(x_{\text{new}} \mid \theta, x_{\text{old}}) p(x_{\text{old}} \mid \theta) p(\theta)}{\sum p(x_{\text{old}} \mid \theta) p(\theta)} \quad (5)$$

$$= \frac{\sum p(x_{\text{new}}, x_{\text{old}} \mid \theta) p(\theta)}{\sum p(x_{\text{old}} \mid \theta) p(\theta)}.$$

Note that the summations are understood to be over $\theta$, and thus the functions are independent of $\theta$—we have removed $\theta$ by marginalization.

For order-independent (or exchangeable) processes, the data are conditionally independent given the model. That is, if the true model were known, then the distribution of future data would be independent of past data. Many common statistical models (normal, binomial, etc.) are exchangeable. For these processes we can make the simplification

$$P(x_{\text{new}} \mid \theta, x_{\text{old}}) = p(x_{\text{new}} \mid \theta) \quad (6)$$

when using Equation 5.

## 4 Abductive Prediction

Bayesian transduction is to be sharply contrasted with *abductive prediction*, or "naive" transduction, which corresponds to the standard scientific method [Peirce]. With abductive prediction, one determines the *best* model based on the previous data (abduction[1]) and then uses this model to make predictions about future data. While this may seem reasonable, there may be no clear best model, and the predictions made by choosing the best one express too much certainty, leading to overconfident predictions. Abductive prediction is shown in Figure 4.

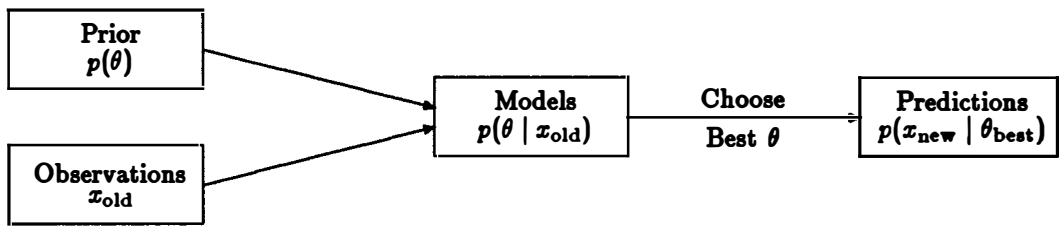

Figure 4: Abductive Prediction

When there is a large amount of prior information, the posterior distribution of the models concentrates nearly all of the weight on a single model, and transduction and abductive prediction will be essentially equivalent. Abduction corresponds to approximating the posterior distribution of the parameters with a delta function at the point estimate of the parameters,

$$p(\theta \mid x_{\text{old}}) \approx \delta_{\theta, \theta_{\text{best}}}. \quad (7)$$

Clearly this approximation is unreasonable unless a single model is strongly indicated. Predictions of AI systems should be valid under all circumstances, since there is typically no check on the validity of built-in approximations.

Referring back to the example of the military commander, abductive prediction corresponds to determining the most likely location of the enemy, and then proceeding as if this were a fact.

Even among those who advocate abduction there is disagreement over how the "best" model should be selected [Peng]. Common approaches include the most probable model based on the posterior distribution of the models (MAP), while others use a mean value or even cruder or subjective methods which are difficult to analyze probabilistically. Using transduction obviates the problem of selecting the best model and recognizes the fallacy of trying to do so. Transduction is only concerned with prediction, so no best model need be selected.

---

[1] We note that Webster defines abduction to be "the unlawful carrying away of a woman for marriage or immoral intercourse."

The difference in methodology between the abduction and transduction methods is due to the differences in their goals. Abduction is concerned with models directly, while transduction is concerned with predictions of data, the models being used only indirectly. Since in a real system only the data are observable, not the models, we feel that direct inference about the models is unneccesary. For example, if we were to replace the military commander with an expert system (May this never happen!), we would judge its performance by its ability to defeat its enemy, which is directly related to its ability to predict its enemy and the results of its own actions. We would not judge the accuracy of its internal models of the world except in that they lead to accurate predictions and thus good behaviour.

## 4.1 Moments of Predictive Distributions

For models with adjustable parameters, we often wish to know only the *moments* of the posterior predictive distribution. For example, we might only wish to know the *mean* number of defective widgets that a machine will produce, rather than the complete distribution. The moments of the predictive distribution, $p(x_{\text{new}} \mid x_{\text{old}})$, can be expressed in terms of the moments of the posterior distribution of the parameters, $p(\theta \mid x_{\text{old}})$, and this relationship clarifies the difference between abduction and transduction. In this section we will assume that we are considering a certain family of models which are characterised by a set of parameters. We will henceforth denote these parameters by $\theta$, since the values of the parameters determine the model under consideration. These parameters can be either discrete or continuous.

The moments of the predictive distribution can be found by expressing the moments of the likelihood function in terms of the parameters, $\theta$, and then finding the moments of the posterior distribution of these functions of $\theta$ given $x_{\text{old}}$. For example, the mean number of defective widgets can be expressed in terms of the mean of the posterior distribution of $p$, the unknown proportion of defective widgets. We find that the first two moments—the expectation, E, and variance, V—are

$$
\begin{aligned}
E(x_{\text{new}} \mid x_{\text{old}}) &= E(E(x_{\text{new}} \mid \theta, x_{\text{old}}) \mid x_{\text{old}}) \\
V(x_{\text{new}} \mid x_{\text{old}}) &= E(V(x_{\text{new}} \mid \theta, x_{\text{old}}) \mid x_{\text{old}}) + V(E(x_{\text{new}} \mid \theta, x_{\text{old}}) \mid x_{\text{old}}).
\end{aligned}
\tag{8}
$$

Again, for exchangeable processes we can make the simplification as in Equation 6,

$$E(x_{\text{new}} \mid \theta, x_{\text{old}}) = E(x_{\text{new}} \mid \theta). \tag{9}$$

It is important to understand that $E(x_{\text{new}} \mid \theta)$ is simply some function of $\theta$ and that this function has an expectation based on the posterior distribution of $\theta$ given $x_{\text{old}}$.

As an example of the use of moments, consider an independent normal process. In this case the model parameters are the mean and variance of the process, $\theta = (\hat{x}, \sigma^2)$. The data consist of the set of real values, x, which are observed. For the normal distribution, $E(\mathbf{x} \mid \hat{x}, \sigma^2) = \hat{x}$, and $V(\mathbf{x} \mid \hat{x}, \sigma^2) = \sigma^2$. That is, the two parameters of the normal model *are themselves* the first two moments of the likelihood function.

Applying Equation 8 we find that the moments of the predictive distribution, $p(\mathbf{x}_{\text{new}} \mid \mathbf{x}_{\text{old}})$, are

$$
\begin{aligned}
E(\mathbf{x}_{\text{new}} \mid \mathbf{x}_{\text{old}}) &= E(\hat{x} \mid \mathbf{x}_{\text{old}}), \\
V(\mathbf{x}_{\text{new}} \mid \mathbf{x}_{\text{old}}) &= E(\sigma^2 \mid \mathbf{x}_{\text{old}}) + V(\hat{x} \mid \mathbf{x}_{\text{old}}).
\end{aligned}
\tag{10}
$$

Thus the predicted mean of future data is the same as the mean of the posterior distribution of $\hat{x}$ given the old data, as one might expect. However, the variance of future data is *not* simply the mean of the posterior distribution of $\sigma^2$, but additionally includes the variance of the posterior distribution of $\hat{x}$. For very large prior samples the variance of the posterior distribution of the parameters is very small so this second term become less important.

It is fundamentally important to understand that there are two sources of uncertainty (variance) in predictions about future data corresponding to the two terms of Equation 8. The first term is due to the fact that even when the parameters, $\theta$, are known *perfectly*, the model predicts a distribution with a particular variance. The second term is due to the fact that the parameters are *not* known with certainty, and our estimate of the mean of the likelihood function has a certain variance, which also contributes to the predicted variance of future data. Equation 8 shows that these two terms are additive. Further, if the data are distributed *independently* of each other in the likelihood function, they need *not* be distributed independently in the predictive distribution. The correlation is due to the second term of the variance of the predictive distribution (Equation 8).

## 5 Example

Consider an exercise taken from *Introduction to Mathematical Statistics* [Hoel, p. 95, ex. 69]:

> *A manufacturer of cotter pins knows from experience that 6 percent of his product is defective. If he sells pins in boxes of 100 and guarantees that at most 10 pins will be defective, what is the probability that a box will fail to meet the guaranteed quality?*

Clearly the author is expecting us to assume that the defects are to be modelled by a binomial process in which the proportion of defective pins, $p$, is known to be 0.06. Thus the number of defective pins, $r$, in a sample of $n$ pins is given by the binomial likelihood function,

$$p(r \mid n, p) = \frac{n!}{r!(n-r)!} p^r (1-p)^{n-r}. \tag{11}$$

Let us examine what this procedure is implying. The author would lead us to believe that the actual proportion of defective pins can be observed directly—that is, determined from experience. Since the manufacturer of the pins can only have observed a finite sample, he can only ever know this proportion to a certain accuracy. The posterior distribution of the proportion of defective pins based on observing $r_0$ defective pins in a sample of $n_0$, is given by the beta distribution [Jaynes],

$$p(p \mid r_0, n_0) = \frac{(n_0 - 1)!}{(r_0 - 1)!(n_0 - r_0 - 1)!} p^{r_0 - 1} (1-p)^{n_0 - r_0 - 1} \, dp. \tag{12}$$

This posterior distribution was derived using the non-informative prior distribution of $p$ [Jaynes],

$$p(p) \propto p^{-1}(1-p)^{-1} \, dp. \tag{13}$$

If this posterior distribution is now combined with the binomial likelihood function and the parameter $p$ is integrated out as described in section 3.3, one finds that the distribution of the number of defective pins, $r$, in a sample of $n$ pins based on the observation of $r_0$ defective pins in a previous sample of $n_0$ is the beta-binomial distribution [Raiffa, p. 237],

$$p(r \mid n, r_0, n_0) = \frac{n!}{r!(n-r)!} \frac{(r_0 + r - 1)!}{(r_0 - 1)!} \frac{(n_0 - r_0 + n - r - 1)!}{(n_0 - r_0 - 1)!} \frac{(n_0 - 1)!}{(n_0 + n - 1)!}. \tag{14}$$

Figure 5 shows a comparison between the predictions of the beta-binomial distribution (with various prior sample sizes, $n_0$, and with $r_0/n_0 = 0.06$) and those made from the binomial distribution (with $p = 0.06$). The first column shows the prior sample size, $n_0$. The bottom row shows the results for the binomial distribution which is identical to the limiting case of the beta-binomial distribution where $n_0 \to \infty$. The second and third columns show the predicted mean and variance of the proportion of defects in the next 100 samples. The fourth column shows the proportion of boxes which can be expected to be defective based on the previous sample, $p(r > 10 \mid n = 100, r_0/n_0 = 0.06)$. The last column shows the percentage of additional boxes which cab be expected to be defective as compared to the predictions using the binomial distribution (bottom row). That is, the last column shows the percentage error of the binomial approximation to the beta-binomial distribution, and thus is a measure of the overconfidence of this approximation.

It is worth noting that accurate estimation of the variance of the future samples is as important as estimation of the mean when evaluating questions involving cumulative probabilities (tail areas). The marked increase in the estimated proportion of boxes rejected using the predictive distribution is due directly to its higher variance. Also, the accuracy of the binomial approximation is not simply a function of the size of the prior sample, but rather the ratio of the size of the sample one is trying to predict to the size of the initial sample. Thus, even if one million initial samples have been observed, predictions about the next several million samples will be inaccurate if the binomial approximation is used.

| Prior Sample Size | Mean of Pin Defects | Standard Deviation of Defects | Percentage of Boxes Rejected | Additional Boxes Rejected |
|---|---|---|---|---|
| 100 | 6.0 % | 3.342 % | 9.922 % | 163.8 % |
| 1000 | 6.0 % | 2.490 % | 4.525 % | 20.32 % |
| 10,000 | 6.0 % | 2.387 % | 3.838 % | 2.061 % |
| 100,000 | 6.0 % | 2.376 % | 3.768 % | .2063 % |
| $\infty$ | 6.0 % | 2.375 % | 3.760 % | 0 % |

Figure 5: Comparison of Predictions of Rejected Boxes

A statistician would never try to use the binomial method for small initial samples, but in an AI system there will be no one to inject this common sense. If we are to build artificially intelligent systems, we must either provide rules to supply this common sense, or simply build in inference methods which are valid in both the small and large sample limits in the first place. We advocate the latter procedure for AI systems.

To better understand the behaviour of the predictive distribution, $p(x_{\text{new}} \mid x_{\text{old}})$, we consider some special cases. For predicting only the next observation, $n = 1$, the predictive distribution, $p(r \mid n = 1, r_0, n_0)$, is equal to the likelihood function with $p = r_0/n_0$, i.e. $p(r \mid n = 1, p = r_0/n_0)$,

$$p(r \mid n = 1, r_0, n_0) = p(r \mid n = 1, p = r_0/n_0) = \begin{cases} \frac{r_0}{n_0}, & r = 1, \\ 1 - \frac{r_0}{n_0}, & r = 0. \end{cases} \quad (15)$$

This fact is peculiar to the binomial process. For example, the predictive distribution of the next sample from a normal process, given an initial sample, is *not* normal for *any* choice of the mean and variance. This peculiarity of the binomial model may be the cause of much of the historical confusion between probabilities and frequencies.

Equation 15 shows that if we are only interested in making predictions about the *next* sample, then we can use the binomial likelihood with $p = r_0/n_0$. If we apply this procedure recursively, predicting one additional sample after each step (updating the prior sample at each step) we find that the probability of observing one defect out of the next two samples is

$$\begin{aligned} p(r = 1 \mid n = 2, r_0, n_0) &= p(1 \mid 1, r_0, n_0) \, p(0 \mid 1, r_0 + 1, n_0 + 1) + p(0 \mid 1, r_0, n_0) \, p(1 \mid 1, r_0, n_0 + 1) \\ &= \frac{r_0}{n_0} \left(1 - \frac{r_0 + 1}{n_0 + 1}\right) + \left(1 - \frac{r_0}{n_0}\right) \frac{r_0}{n_0 + 1} = 2 \frac{r_0(n_0 - r_0)}{n_0(n_0 + 1)}. \end{aligned} \quad (16)$$

This process is simply calculating the beta-binomial distribution (Equation 14) sequentially.

Let us now investigate the moments of the likelihood function and predictive distribution. The first two moments of the future proportion, $r/n$, using the binomial likelihood function (Equation 11) are

$$E\left(\frac{r}{n} \mid n, p\right) = p, \qquad V\left(\frac{r}{n} \mid n, p\right) = \frac{1}{n} p(1-p). \quad (17)$$

Note how the variance is asymptotically proportional to $1/n$. Thus if we know $p$ with *absolute certainty* the variance of $r/n$ will tend to zero for large enough future samples.

To calculate the moments of $r/n$ given the predictive beta-binomial distribution we first calculate the moments of $p$ and $p(1-p)/n$ given the posterior distribution (Equation 12). These are then combined as

per Equation 8 to give

$$E\left(\frac{r}{n}\bigg|n,r_0,n_0\right) = \frac{r_0}{n_0}, \qquad V\left(\frac{r}{n}\bigg|n,r_0,n_0\right) = \left(\frac{1}{n}+\frac{1}{n_0}\right)\left(\frac{n_0}{n_0+1}\right)\left(\frac{r_0}{n_0}\right)\left[1-\left(\frac{r_0}{n_0}\right)\right]. \qquad (18)$$

The predicted mean is simply $r_0/n_0$, as one expects. Although the predicted variance has a $1/n$ term which is similar to the expression for the variance of the likelihood function (with $p = r_0/n_0$), it also has an additional $1/n_0$ term which is dependent on the *future* sample size. An important difference between the variance of the likelihood function (Equation 17) and that of the predictive distribution (Equation 18) is in their asymptotic behaviour as $n \to \infty$. The variance of the binomial distribution tends to zero, while that of the beta-binomial tends to the posterior variance of $p$. Thus, although the averaging effect of many future samples tends to decrease the variance of $r/n$ in large future samples, this variance can never be decreased below the variance of $p$ given by the previous sample. In other words, the predictions cannot converge absolutely to the *actual* proportion if this proportion is not *actually* known!

## 6 Model Revision

In Figure 3, it is assumed that the correct class of models (family of distributions) has been selected. The resulting predictive distribution, $p(x_{\text{old}}, x_{\text{new}})$, is implicitly conditioned by the model class. That is, although the parameters have been integrated out, we are still only considering models of that parameterised family. For example, consider a normal model where one expects the density of observed data to have the familiar normal distribution. If the data set includes some "glitch" points (outliers) in the tails of the normal distribution, then use of the normal model will give poor predictions. Outliers often arise in real data due to mis-reading of instruments, corruption of recordings and the like. Induction and transduction can easily be extended to such situations.

One method is to use an appropriate extended model in the first place. In the case of the normal model with outliers, where the possibility of outliers is known in advance, a suitable mixed model should be used from the beginning. In such a mixed model, the likelihood allows each data point to be accounted for by either a normal distribution or a flat distribution that reflects the expected distribution of possible outliers. The mixed model also should include a parameter that gives the probability of outliers. However, in many situations, it is impossible or impractical to include all the possible model deviations in one grand mixed model. In such cases, it is a good heuristic to first detect when the current model is breaking down and then revise the model appropriately. This revised model is then used as if it had been the model from the beginning, and the old model is discarded.

## 7 Discussion

A major point of this paper is that Bayesian predictive inference involves a mapping from the current data to future data via a model space. That is, in the final prediction, the model has been integrated out, and thus does not appear explicitly in the prediction, as shown in Figure 3. This implies that finding the "best" model with the aim of using it for prediction, as is typical in scientific discovery, does not lead to optimal prediction. In practice, if there is sufficient initial information to make one model overwhelmingly likely, then the difference between choosing the best model and summing over all possible models is very slight. However, for AI it is preferable to have a theory that works regardless of how much initial information is available.

The elimination of the model in transduction is echoed in philosophy of science debates. Many philosophers [e.g. Pearson] believe in *phenomenalism*—the view that the aim of science is to discover regular patterns among our sensations that enable us to predict future sensations. This is closely related to the idea of *instrumentalism*—the view that scientific theories that make reference to unobservable entities (e.g. neutrons, emotions, etc.) are not literally true; rather, they should be regarded as useful instruments for dealing with observable events, objects, and properties [Bridgman]. Phenomenalism is largely discredited in the philosophy of science community [Salmon], yet this paper shows that Bayesian inference supports phenomenalism.

Part of the explanation is that the aim of scientific investigation is to discover the "true" model of the world—to "explain" it— not just to make predictions. Scientists do not doubt the reality of such theoretical constructs as electrons and quarks, even though they are fundamentally unobservable with human senses. For scientists, observation has so strongly pinned down a specific model (theory) that they no longer doubt the existence of the theoretical constructs. Last century, the atomic theory of matter was regarded as "just a theory" and there were doubts about the existence of atoms. Theoretical constructs, such as cosmic strings, neutrino mass, etc., currently enjoy a similar existential status somewhere between fairies and photons. It seems that at some stage of the development of theories, a particular theory becomes the dominant paradigm and the previously competing theories become historical footnotes [e.g. Kuhn]. Of course, there is never a stage at which one theory becomes right and all the others wrong—what happens is that the dominant theory becomes overwhelmingly more likely than its competitors. Those familiar with the see-saw acceptance of the particle versus wave theories of light know that belief in a particular theory as absolutely correct is unsound.

In summary, the scientific practice of assuming that the current best theory is correct and basing all predictions on this theory, is a good heuristic approach. It avoids having to explicitly calculate the predictions of all possible competing theories. However, if the goal is to make as accurate a prediction of future events as possible, the transduction approach discussed in this paper is optimal. For AI decision making problems, transduction is the method of choice, when the cost of doing transduction relative to heuristic alternatives is lower than the cost of a wrong decision.

## 8 Conclusion

This paper has presented the correct Bayesian method for making predictions from current data to future data via a class of models. This method we call *transduction*. Transduction requires that prediction be based on averaging over all the possible models, and not just selecting the best model. The scientific method and naive statistical approaches incorrectly select the best model for predictions. A numerical example shows that this abductive approach can lead to serious errors in prediction. For AI systems, we argue that the transduction approach should be adopted so that the system will make good predictions regardless of the data on which they are based.